\documentclass[conference]{IEEEtran}
\IEEEoverridecommandlockouts
\usepackage{cite}
\usepackage{amsmath,amssymb,amsfonts}
\usepackage{algorithmic}
\usepackage{graphicx}
\usepackage{textcomp}
\usepackage{xcolor}
\usepackage{hyperref}
\hypersetup{
	colorlinks=true,
	linkcolor=blue,
    filecolor=magenta, 
    urlcolor=cyan
}
\def\BibTeX{{\rm B\kern-.05em{\sc i\kern-.025em b}\kern-.08em
    T\kern-.1667em\lower.7ex\hbox{E}\kern-.125emX}}
\begin{document}

\title{Outlier Detection and Spatial Analysis Algorithms}

\author{\IEEEauthorblockN{Jacob John}
\IEEEauthorblockA{\textit{School of Computer Science and Engineering (SCOPE)} \\
\textit{Vellore Institute of Technology}\\
Vellore, India \\
jacob.john2016@vitalum.ac.in}
}

\maketitle

\begin{abstract}
Outlier detection is a significant area in data mining. It can be either used to pre-process the data prior to an analysis or post the processing phase (before visualization) depending on the effectiveness of the outlier and its importance. \cite{b10} Outlier detection extends to several fields such as detection of credit card fraud, network intrusions, machine failure prediction, potential terrorist attacks and so on. Outliers are those data points with characteristics considerably different. They deviate from the data set causing inconsistencies, noise and anomalies during analysis and result in modification of the original points. \cite{b5} However, a common misconception is that outliers have to be immediately eliminated or replaced from the data set. Such points could be considered useful if analyzed separately as they could be obtained from a separate mechanism entirely making it important to the research question. This study surveys the different methods of outlier detection for spatial analysis. Spatial data or geospatial data are those that exhibit geographic properties or attributes such as position or areas. An example would be weather data such as precipitation, temperature, wind velocity and so on collected for a defined region.
\end{abstract}


\section{Literature Survey}
Success of anomaly detection for spatial mining techniques relies on the definition of the neighborhood which sets the frame of reference. Spatial autocorrelation says that objects that are in proximity to each other tend to behave similarly. Spatial heterogeneity, on the other hand, focuses on the undergoing processes and causes the data within zones to be distinct or uncorrelated. 

This paper \cite{b9} highlights the limitations of spatial autocorrelation and identifies the importance of spatial heterogeneity coupled with the creation of data sets. It also collates data from water sensors. Furthermore, by identifying abnormally behaving water monitoring sensors, it shows the importance of spatial heterogeneity by evaluating the impact after applying outlier detection and weighing the importance of spatial heterogeneity. Thus, concluding that anomalous behavior of objects can be recorded only when both spatial autocorrelation and heterogeneity are considered, in order to generate neighborhoods (both micro and macro). In addition to this, it aids in retrieving or alleviating (i) spatio-temporal outliers, (ii) spatial outliers and (iii) spatio-temporally coalesced outliers.

Take an example of sensor readings collected. These measure the toxicity levels of the water body. The goal is to identify any anomalous data while obtaining water toxicity levels. 

\begin{figure}[htbp]
\centerline{\includegraphics[width=0.4\textwidth]{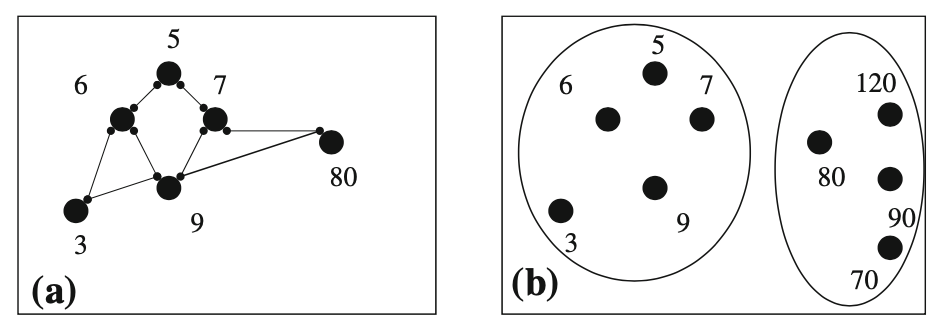}}
\caption{(a): Neighborhood formation only on the basis of spatial autocorrelation (b) Formation taking spatial heterogeneity into consideration.\cite{b9}}
\label{fig1}
\end{figure}

Given in Fig. \ref{fig1} (a), the neighborhood formation is on the basis of spatial autocorrelation. However, note with the value 80 is also a part due to its proximity to the sensors. Therefore, considering relationships obtaining using adjacency leads to frivolous anomalies such as node with the value 80 being included in the set. However, Fig. \ref{fig1} (b) is generated using spatial heterogeneity and leads to the anomaly being segregated in to a new neighborhood or cluster. Thereby, emphasizing the importance of spatial and inferential relationships.

This study \cite{b14} underlines the different approaches for unsupervised outlier detection in a heurisitic manner using data from the US Census. The approaches have been categorized as global – based on the complete data set and local – based on selected data objects. Furthermore, the paper recognizes the need for mentioning which ``degree and notion of locality" with clarity when proposing a new method. This is because of its numerous perspectives on locality. 

\begin{figure}[htbp]
\centerline{\includegraphics[width=0.4\textwidth]{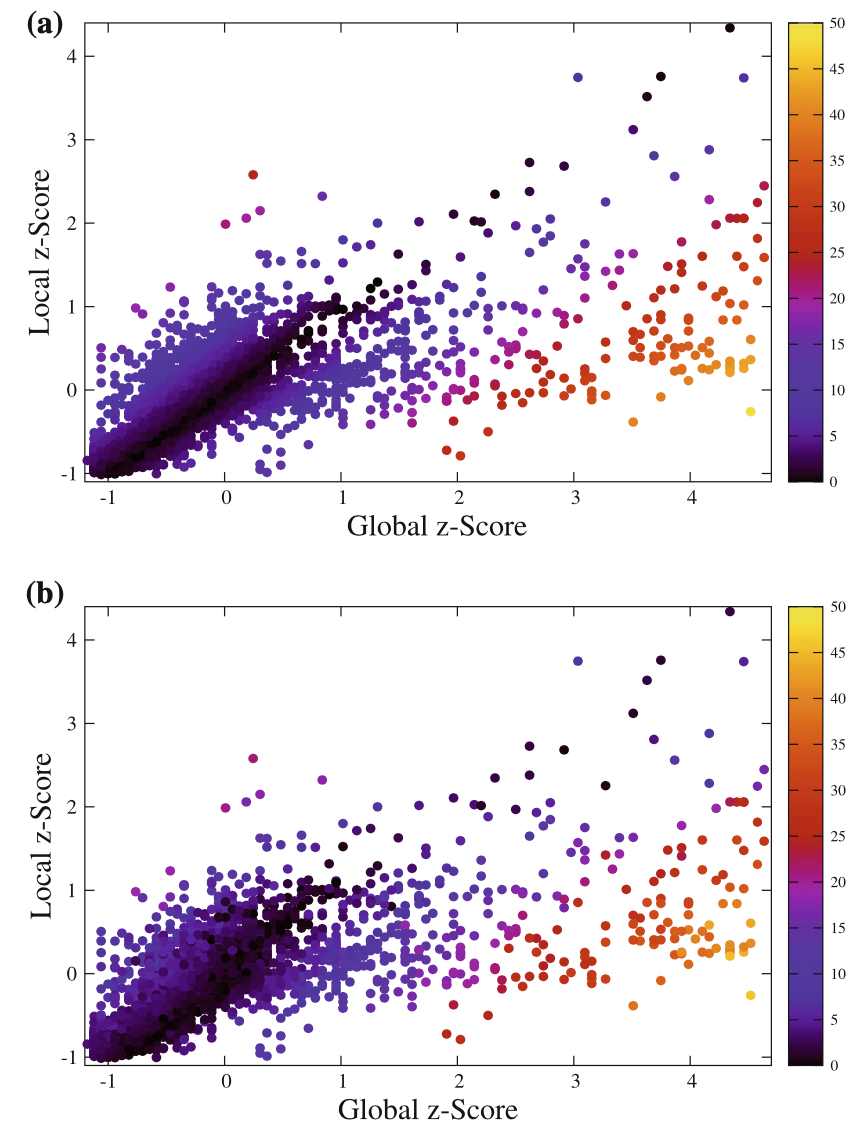}}
\caption{Moran Scatterplot with outlier scores for US Census land use data—basic methods. a Z-statistic, b median. \cite{b14}}
\label{fig2}
\end{figure}

Spatial neighborhood is a special case of locality. The study mentions in order to analyze spatial attributes, $k$ nearest neighbor algorithm are typically due to the data's uniform size. Fig. \ref{fig2} (a) and (b) are visualizations curated using statistical approaches. Figure (a) uses Z-statistic for identifying outliers and (b) uses the median score.

This study \cite{b12} elucidates and illustrates the application of data cleaners using an exemplified approach. Data cleaners are used to annihilate `patchy' and `isolated' outliers using a robust algorithm described below. Its purpose further extends to the `cleaning' of spatial lattice data with method that uses a find-and-replace form of approach.

The class of data cleaners identified by this paper works as given in \eqref{eq1}.

\begin{equation}
\psi(y_{u,v};M,g^0,g^1)=\begin{cases}
y_{u,v}\ \text{if}\ |y_{u,v} - g^0(y_{u,v})| \le M\\
g^1(y_{u,v})\ \text{otherwise}\\
\end{cases}\label{eq1}
\end{equation}

Where, $y_{u,v}$ is an observation. $y_{u,v}$ is considered an outlier if $|y_{u,v} - g^0(y_{u,v})| \le M$. The data cleaner algorithm replaces the outlier with $g^1(y_{u,v})$ when $y_{u,v}$ is an outlier while the rest of the data is unaffected. ``A sensible choice for $g^0$ is the median of the observed values." Thus $|y_{u,v} - g^0(y_{u,v})|$ can also be referred to as the deviation from the median in this case.

A value for $g^1$ in this case, can be used as the median to smoothen the data after removing the outliers. However, the median in this case isn't calculated for the entire lattice (or data set) but rather the nearest four elements to the outlier that is being replaced. This process starts at $y_{1,1}$ and proceeds row-wise until the entire lattice is covered. Furthermore, since the original data is used for calculating averages rather than the lattice after replacing the elements, the ordering of this process doesn't impact the detection. This is method is also referred to as the neighboring median cleaner (NMC).

The diagrams below are the application of NMC with respect to botanic data for the Ceutorhynchus assimilis. 

\begin{figure}[htbp]
\centerline{\includegraphics[width=0.4\textwidth]{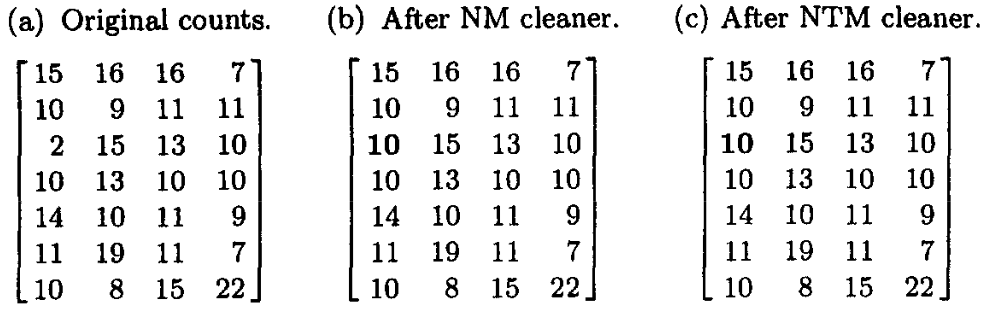}}
\caption{Seed weevil counts before and after cleaning. \cite{b12}}
\label{fig3}
\end{figure}

\begin{figure}[htbp]
\centerline{\includegraphics[width=0.4\textwidth]{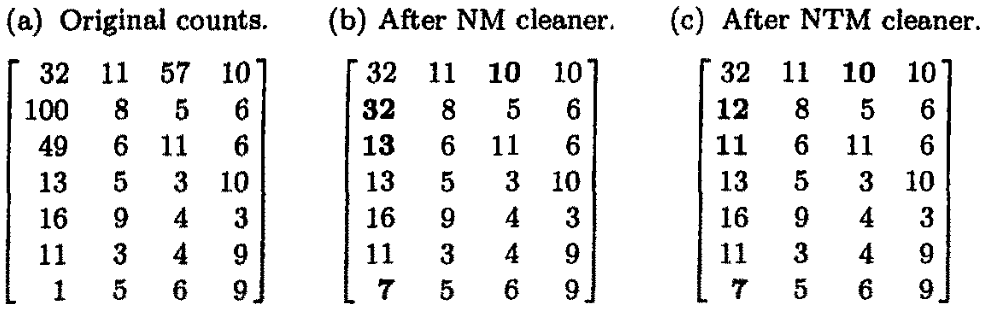}}
\caption{Pollen baetle counts before and after cleaning. \cite{b12}}
\label{fig4}
\end{figure}

\begin{figure}[htbp]
\centerline{\includegraphics[width=0.4\textwidth]{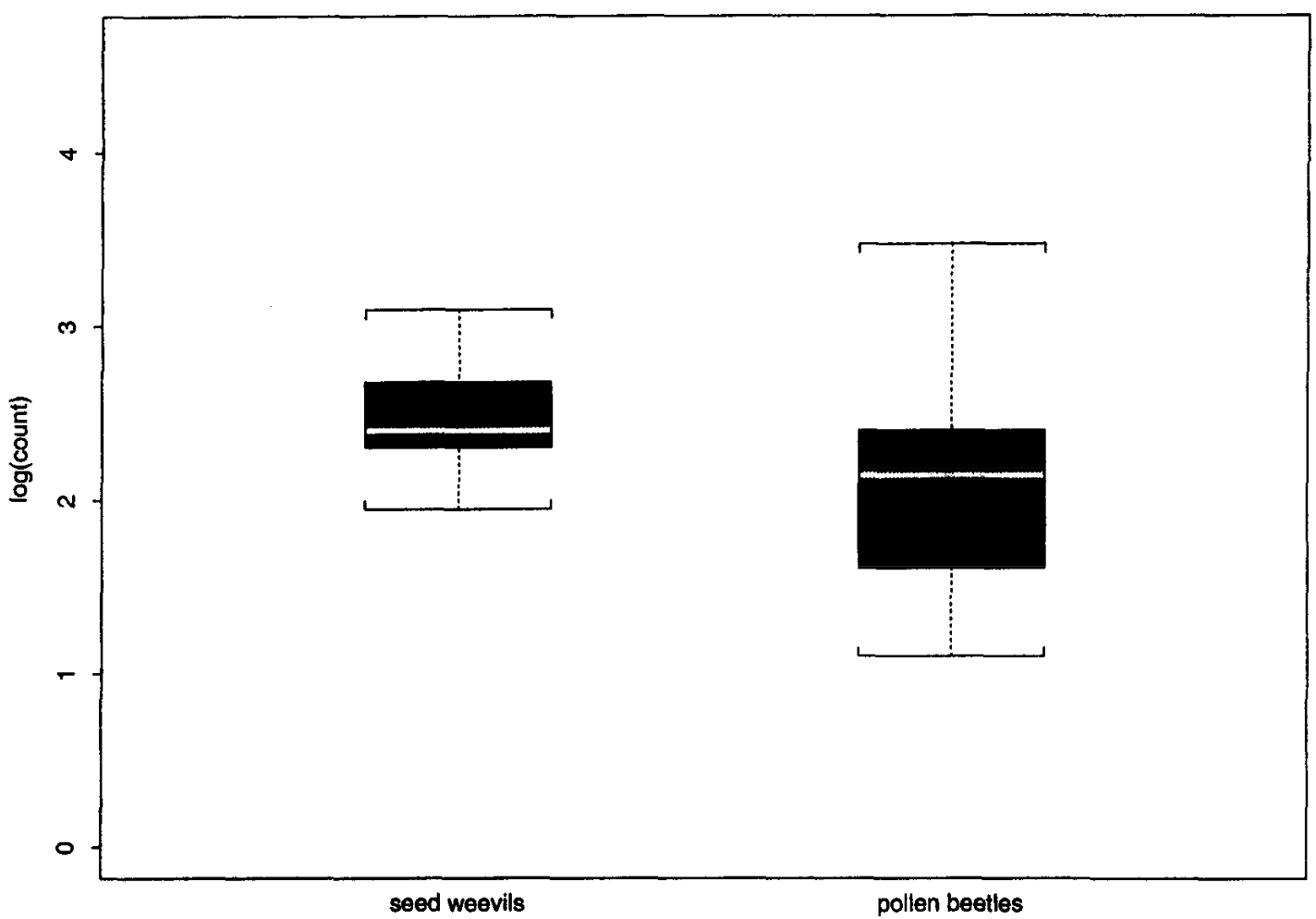}}
\caption{Box and whisker plots of seed weevil and pollen beetles using data from Fig. \ref{fig3} and Fig. \ref{fig4} after log transformation\cite{b12}}
\label{fig5}
\end{figure}

This paper \cite{b11} considers data available from 22 different monitoring stations located in Basse-Normandie and Haute-Normandie regions in France. The data collected is for the pollutant PM10 and are hourly measurements conducted by an official association called Air Normand. The data set available is from January 1, 2013 to May 31, 2013 which means its relatively new (roughly 6 years from the date of writing this exploratory study). Furthermore, the data is used to illustrate and compare two methods namely a non-stationary spatial approach and a stationary approach based on kriging methodology for detection of univariate and multivariate outliers in the data and then extenuating them.

\begin{figure}[htbp]
\centerline{\includegraphics[width=0.4\textwidth]{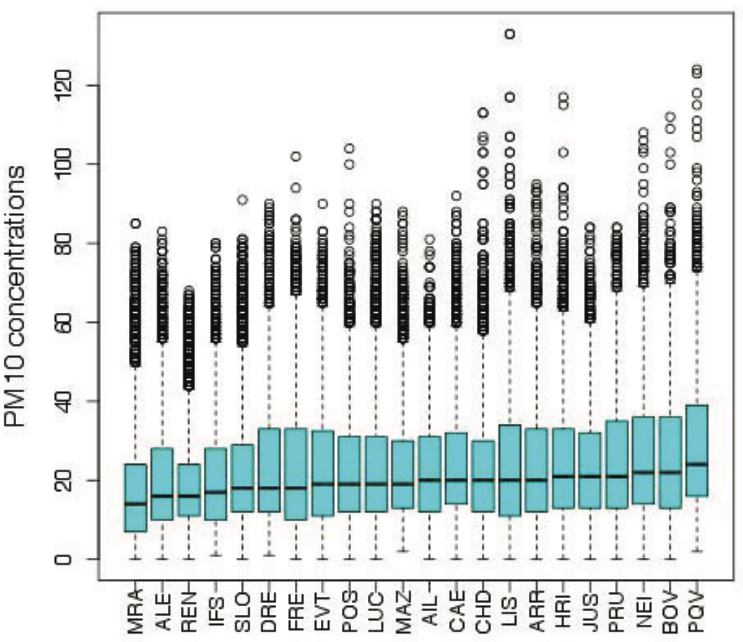}}
\caption{Boxplot obtained of validated PM10 concentrations from 22 different stations\cite{b11}}
\label{fig6}
\end{figure}

Given in figure \ref{fig6} are boxplots of validated PM10 concentrations from Air Normand of the learning set. After observing a relatively homogenous pattern can be inferred as all the medians for the data sets line up within a margin of error.

The first method in identifying outliers and for spatial predictions follows a non-stationary approach as discussed earlier. This uses the \textit{k} nearest neighborhood algorithm and compares concentrations of a given site with the weighted median of neighboring sites. This method is used in a technique referred to as jackknife as according to the paper, which is leaving out observations which are anomalous to the data set to calculate more accurate estimates

The second method on the other hand, follows a stationary spatial approach based on ``kriging the differences between current and reference observations" to give the best linear unbiased prediction of the intermediate observations.

This paper \cite{b13} addresses the issues related to multivariate outlier detection using unsupervised learning. Introduced by Kohonen, the self-organized map (SOM) is an artificial neural network (ANN) for dimensionality reduction and clustering of high-dimensionality data. Tools such as median interneuron distance matrix further develops estimates in conjunction with Sammon’s mapping algorithms. These also help in identifying the ``first type of outlyingness". The second level is formed using SOM quantization errors (QEs). In addition, this paper infers after analyzing a number of techniques, based on other trained SOMs, that they work well in collaboration with each other. 

Outliers degrade classical principles and cluster analysis by causing discrepancies in cluster center and inflating or deflating variances and hence deteriorating the quality and accuracy of the results obtained. Furthermore, this paper suggests that multivariate outliers are relatively hard to detect and cause the masking and swamping effect and could possibility increase the dimensionality of the data. SOMs, in this case, are relatively fast and inexpensive when the dimensionality of the data is large. Furthermore, this method isn’t model-dependent.

An example \cite{b4} provided is the Local Density-Based Spatial Clustering of Applications with Noise (LDBSCAN) and merges both Density-Based Spatial Clustering of Applications with Noise (DBSCAN) and Local Outlier Factor (LOF). DBSCAN is meant for clustering algorithms. LOF provides outliers with a numerical value or as the papers describes it, ``gives a quantitative measure of outlierness to each object". \cite{b4} It can be inferred that a high LOF leads to a potential discrepant outlier. This method takes into consideration that the multi-granularity deviation factor (MDEF) is a result of the fact that potential outliers are isolated from its neighborhood rather than only the whole dataset. The MDEF copes up with these outlying clusters and helps smoothen the data. 

\begin{figure}[htbp]
\centerline{\includegraphics[width=0.4\textwidth]{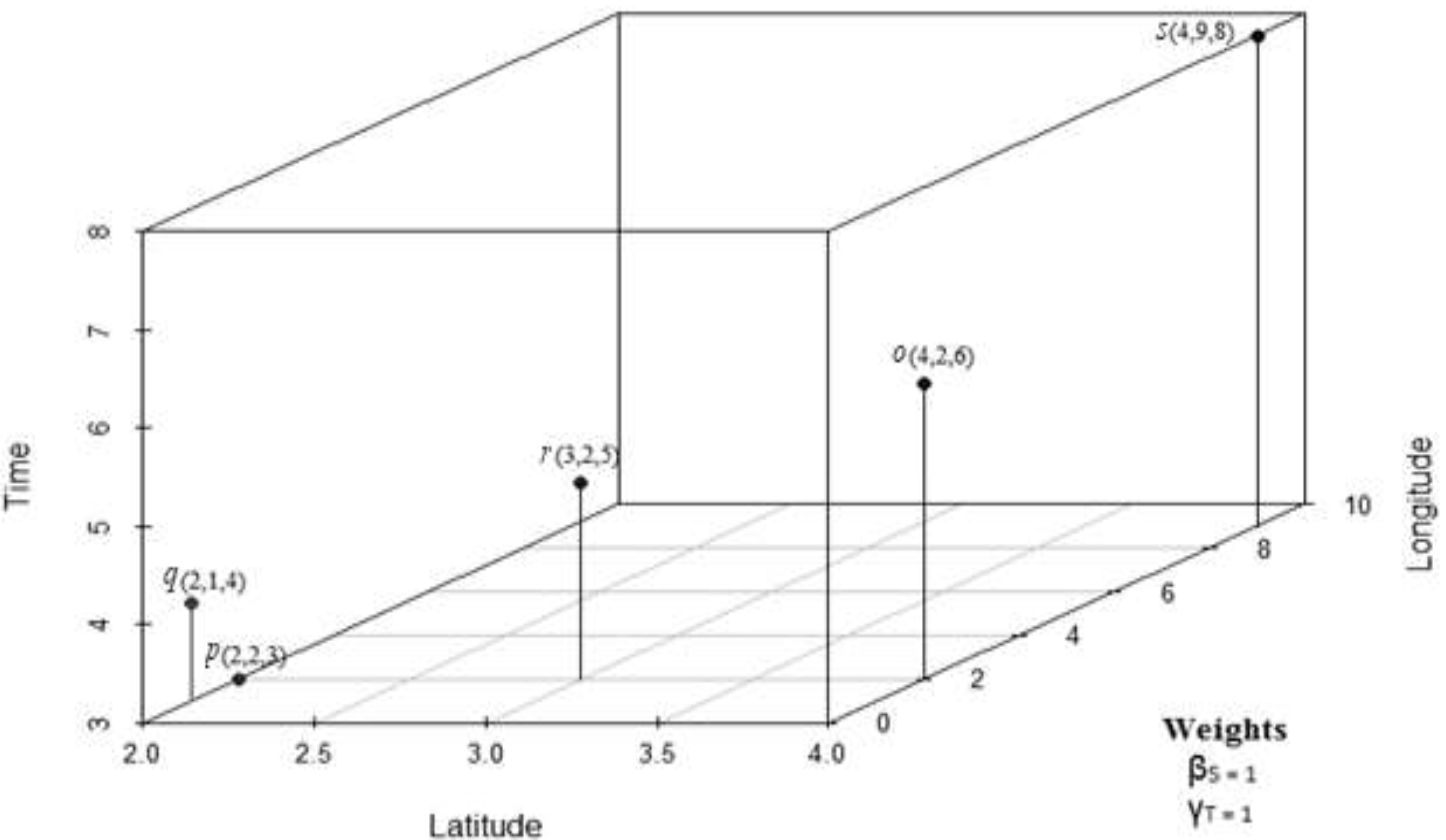}}
\caption{Spatio-Temporal distance of object \textit{p} to object \textit{o} and \textit{q,r,s} along with their respective spatio-temporal attribute values latitude, longitude, time, and the weights.\cite{b4}}
\label{fig7}
\end{figure}

This paper proposes a new method for outlier detection and extends upon the importance of spatial or temporal continuity. The technique proposed is the Spatio-Temporal Behavioral Outlier Factor (ST-BOF). This again is used to measure potential outliers. As compared to LOF that uses all attributes of the data points, ST-BOF only uses the spatial-temporal attributes and behavioral attributes. These are then used to describe the outlierness of the data points and context to define the neighborhood. The definition for ST-BOF is illustrated as in Fig \ref{fig7}.

This paper \cite{b8} associates spatial outliers in wind erosion with severe weather events. In this case, high rate of erosion and sedimentation in this case would be considered as outliers in a spatial context. This paper uses pins established in nested grids to obtain data for soil erosion and sedimentation for geospatial localities. Furthermore, the variability in the pin height is a response variable for the explanatory variables sedimentation and erosion. Lastly, the maximum outlier for soil sediments is underlined as 22 cm.

Data exploration was carried out using the \textit{geoR}, \textit{gstat} and \textit{Sp} packages. This was done in order to normalize data into distributions and eliminate outliers at an early stage. Other visualization techniques used were Normal Q-Q plot, histograms, variogram cloud and variogram modeling.

\begin{figure}[htbp]
\centerline{\includegraphics[width=0.2\textwidth]{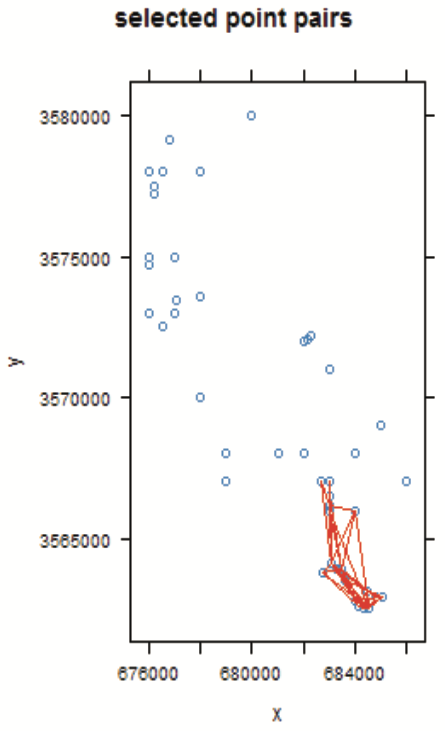}}
\caption{Select point pairs as outliers from the dataset.\cite{b8}}
\label{fig8}
\end{figure}

In conclusion, after considering 47 data points, 13 of these were omitted as anomalous outliers. Furthermore, the paper mentions outliers should still be considered despite its abnormal behavior as they still contain vital and effective information in a spatial analysis. They could also act as special cases. Therefore, they should be assessed as a different set rather than eliminating them completely. Figure \ref{fig8} in this case is one such example where the outliers can be paired and then studied separately.

\begin{figure}[htbp]
\centerline{\includegraphics[width=0.4\textwidth]{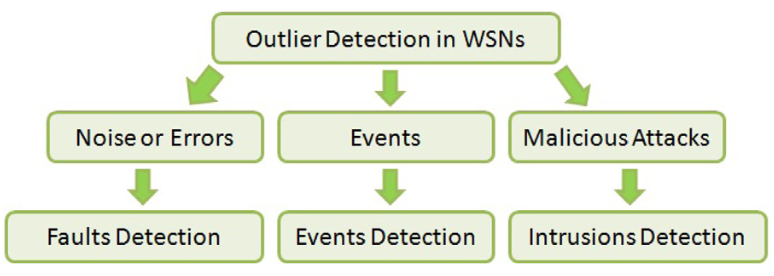}}
\caption{Sources of outliers in a Wireless Sensor Network or WSN.\cite{b1}}
\label{fig9}
\end{figure}

This paper \cite{b1} discusses about wireless sensor networks (WSNs) and how they’ve gained significant attention over the past decade. However, if the density of the network is too high, WSNs are exposed to faults. Henceforth making them vulnerable and allow for malicious attacks to occur. These sensor readings can be inaccurate and unreliable at times and lead to erroneous results. These data points are referred to as outliers, anomalies or abnormal data and cause inconsistency and deviation of the data from the empirical value leading to a lower precision. Figure \ref{fig9} shows the different sources of such outliers in WSNs. This paper outweighs the different methods of outlier detection and evaluates them on characteristics of ``detection mode, architectural structure and correlation extraction." The aim of the paper is to ensure data quality and reliability of data by securely monitoring the data for suspicious activities. 

Statistical based approaches can be successfully implemented provided the right probability distribution model is obtained. They also could encompass temporal attributed to identify outliers. However, these fail when using parametric or non-parametric techniques in real time applications as are only efficient for univariate data and the computational costs are high for multivariate data and real time applications such as WSNs. Artificial Intelligence based approaches are able to ``generalize from limited, noisy and incomplete data.” \cite{b1} But are hard to develop and fine tune. Spatial and temporal semantics make the rules and fuzzy logic more complex and harder to implement and could poses a load on the memory. 

\begin{figure}[htbp]
\centerline{\includegraphics[width=0.2\textwidth]{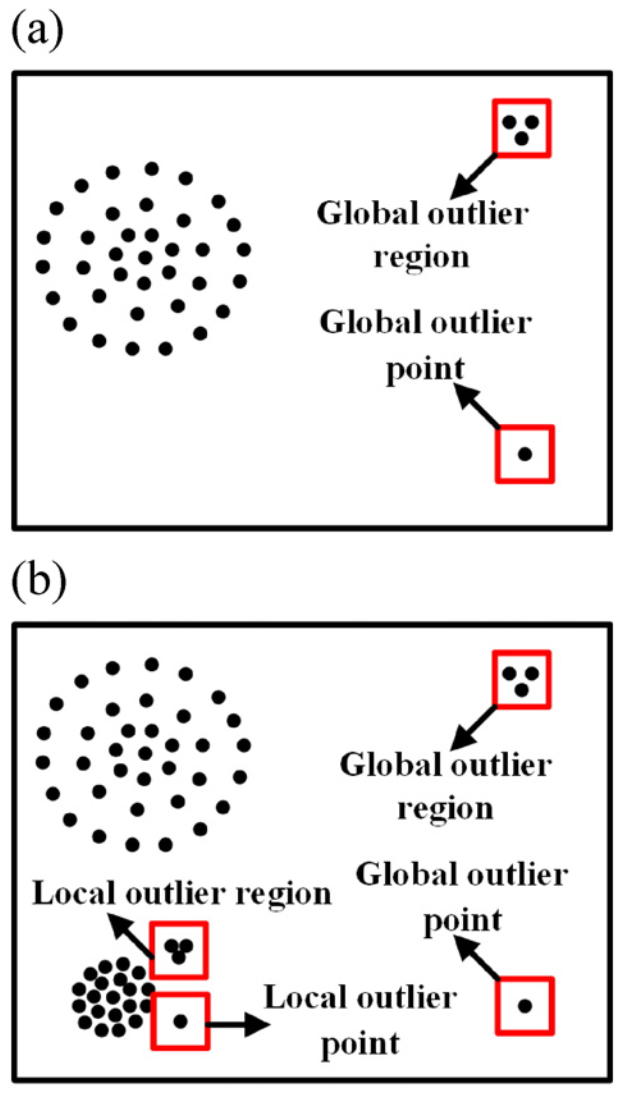}}
\caption{(a) Spatial data for a group of points (b) the Delaunay triangle for the corresponding group.\cite{b16}}
\label{fig10}
\end{figure}

This paper \cite{b16} presents a new adaptive approach for spatial point events outlier detection (SPEOD) using the multilevel constrained Delaunay triangulation method. The first step is establishing the spatial proximity relationships among spatial points events using the Delaunay triangle as shown in Figure \ref{fig10}. This forms the spatial point event data set (SPED). This triangulation method has a time complexity of $\mathcal{O} (N \log N)$.

\begin{figure}[htbp]
\centerline{\includegraphics[width=0.3\textwidth]{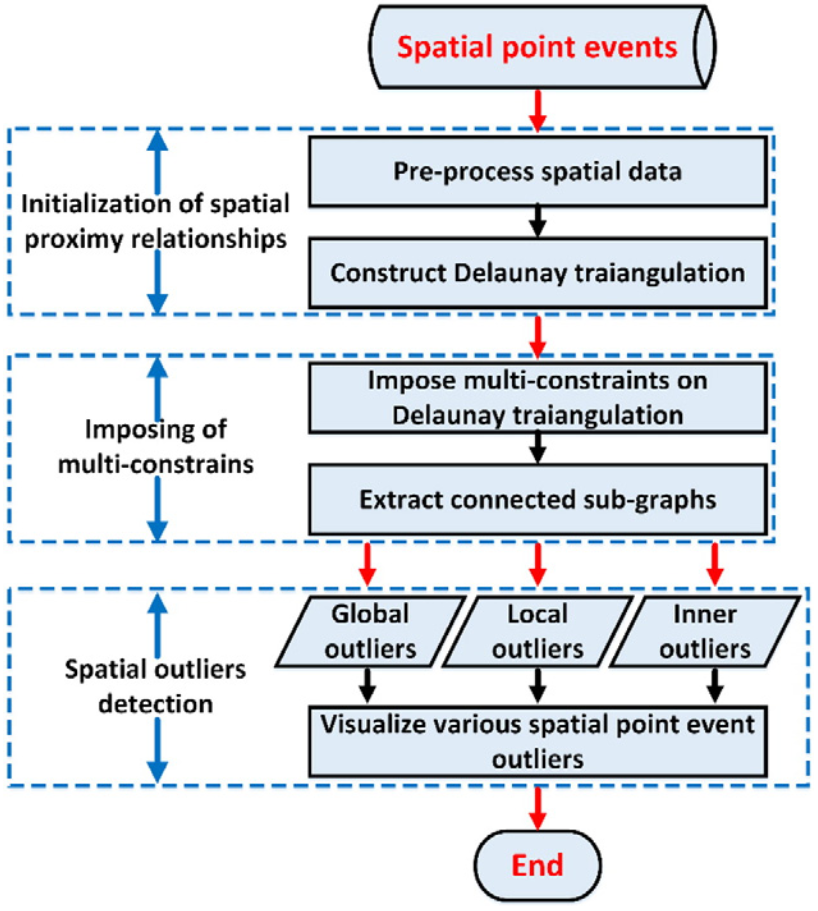}}
\caption{Steps for applying the Delaunay triangulation method for outlier detection.\cite{b16}}
\label{fig11}
\end{figure}

The next step considers statistical characteristics and is used to establish three-level constraints which are described in order to fine tune spatial proximity relationships. Finally, the spatial points create subgraphs by joining the remaining edges amongst events. Those subgraphs with a relatively fewer number of points are referred to as spatial outliers. By gathering those subgraphs and calculating the average volume and corresponding standard deviation can aid in finding spatial outliers. The time complexity for this process is $\mathcal{O}(N)$, where \textit{N} is the number of subgraphs. These steps can be illustrated in Figure \ref{fig11}.

This paper \cite{b19} utilizes the Moran’s ``I index" which can be generally divided into Global Moran’s I index and Local Moran’s I index for spatial autocorrelation. These methods encompass spatial relationships and characteristics of variables and their adjacency between data points. This paper implements these methodologies on the \textit{Geoda} 10.0 software platform for analyzing data for cultivated land. 

Global Moran’s I index as the name suggests, analyzes overall spatial features over a global region and index. The forumal used in this case is as follows:

\begin{equation}
I=\frac{n \times \sum_i^n\sum_j^n w_{ij}(x_i - \bar{x})(x_j - \bar{x})}{(\sum_i^n\sum_j^n w_{ij})\times(\sum_i^n (x_i - \bar{x})^2)}\label{eq2}
\end{equation}

Where in \ref{eq2}, \textit{n} are the number of spatial features, $x_i$ and $x_j$ are the attribute values of feature \textit{i} and \textit{j} respectively,  $\bar{x}$ is the mean of all the features and, $w_{ij}$ is the weight of the spatial matrix and can only take values 0 and 1.

The range of \textit{I} is between 0 and 1 and represents the spatial correlation of the dataset. A Z-test could further be used on the test matrix to obtain the acceptance level of significance.

Local Moran’s I analysis can only establish the degree of spatial correlation for an overall data set but cannot identify whether the correlation is positive or negative for a local region. The formula given is as follows:

\begin{equation}
I_{i} = \frac{x_i - \bar{x}}{S^2_i} \sum_{j=1,j\ne i}^{n} w_{ij}(x_j - \bar{x})\label{eq3}
\end{equation}

Where,

\begin{equation}
S^2_{i} = \frac{\sum^n_{j=1,j \ne 1} (x_i - \bar{x})^2}{n-1} - \bar{x}^2\label{eq4}
\end{equation}

In conclusion, using the K nearest neighbors, the corresponding K values were obtained to be 5, 6 and 5 given that the spatial autocorrelation coefficient are of natural index and the economic index is largest for cultivated land. It shows a global Moran’s index of 0.85 which is spatially positive and has a high aggregation degree. This was done at a 5\% level of significance for statistical hypothesis testing.

\begin{figure}[htbp]
\centerline{\includegraphics[width=0.35\textwidth]{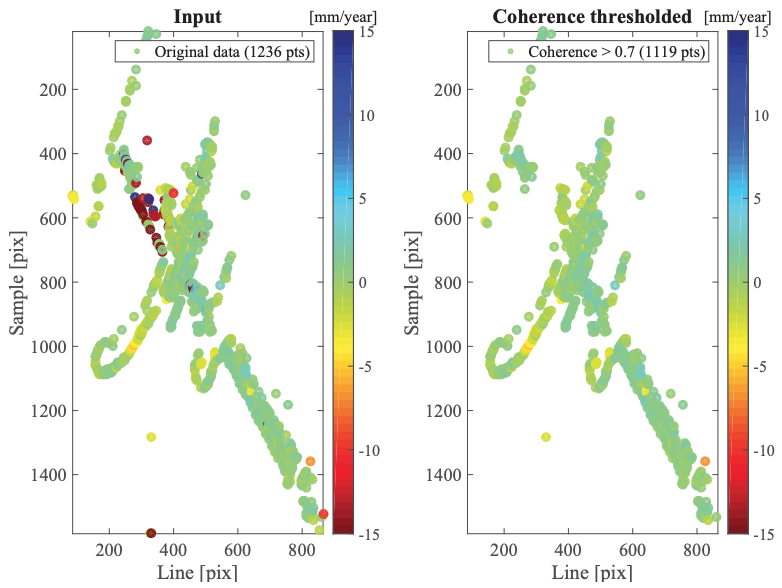}}
\caption{Velocities before and after imposing a threshold of 0.7 on ensemble coherence value. \cite{b2}}
\label{fig12}
\end{figure}

Tested upon data available from Sentinel-1, Multi-temporal InSAR (MT) results show a new approach for detecting multivariate outliers with respect to post-processing of data. This approach \cite{b2} demonstrated an expanding point density by a half of the aggregate sum of standard PS points. The MTI technique overcomes the challenge exhibited by the presence of spatial dependencies among observations. Low coherence areas show enhanced details of deformations, noise and imperfections in this model. MTI is particularly useful when the coherence $\in [0,1]$ and is exceeds the threshold of 0.7. Furthermore, parameters such as velocity and height where used to perform these estimates. Figure \ref{fig12} shows the impact of MTI over the given threshold while maintaining coherence over 0.7.

This paper \cite{b3} studies and presents Self-organizing feature maps (SOFM) as a novel approach for outlier detection for non-spatial attributes. This overcomes the inefficiencies of clustering algorithms, which are being sensitive to input data and unreliable outcomes while processing noise. Furthermore, SOFMs also allows the analysis of multi-dimensional data as well. 

This paper analyzes the features of the SOFMs and the inadequacies are overcome. This is done by implementing the SOFM in an experiment by organizing it in a hierarchical structure using the standard Iris data set. ``The data features include a 3-dimensional data with 500 points and 5 clusters, 4 rational clusters in projection on 2-dimension surface." An extension of clustering theory and SOFM is proposed by this paper called topological similarity. This is complemented by a mathematical model to accompany this similarity theory. This is called the topological similarity matrix which uses adjacency as parameter. 

This paper \cite{b20} defines a spatial-temporal outlier as an ``object whose non-spatial attribute value is significantly different from those of other objects in its spatial and temporal neighbors." This paper uses a Hadoop based approach to evaluate and detect outliers with a proposed algorithm claiming that Hadoop could improve performance of the running algorithm. In order to experiment this, the Ningbo sea tide data set is used to test the validity of this claim.

This algorithm proposed uses the Hadoop Distributed File System (HDFS) and MapReduce which are relatively complex and its composed of four sequential-modular MapReduce programs. MapReduce uses the concept of divide and conquer for a parallel computing program. MapReduce in this case, uses \textit{K} neighbors for outlierness of every spatial location. Furthermore, it’s also for identifying temporal outliers by obtaining the coordinates and non-spatial attributes of the outliers from the original data set. It also arranges the outliers in ascending order.

\begin{figure}[htbp]
\centerline{\includegraphics[width=0.35\textwidth]{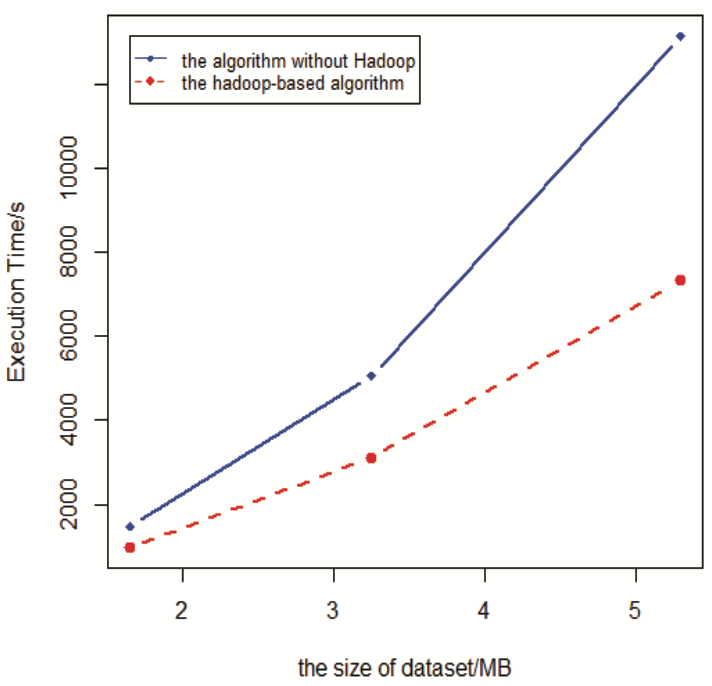}}
\caption{The comparison of the Hadoop-based algorithm and the algorithm without Hadoop. \cite{b20}}
\label{fig13}
\end{figure}

Despite, its complexity, Hadoop performed much better than the approach without. Figure \ref{fig13} illustrates this comparison. 

This paper \cite{b7} discusses the four types of outliers in a geographical setting. Figure 14 illustrates these four different types of settings. Outliers in this case are measured and geographically weighted over Mahalanobis distances or the distance between any two points P and Q and signifies how many standard deviations away is P from the mean of Q. These distances are used to transform data spaces in order to detect outliers.  The data set used to assess the algorithm is simulated data bundled with freshwater chemistry data collected all over Great Britain. 

\begin{figure}[htbp]
\centerline{\includegraphics[width=0.4\textwidth]{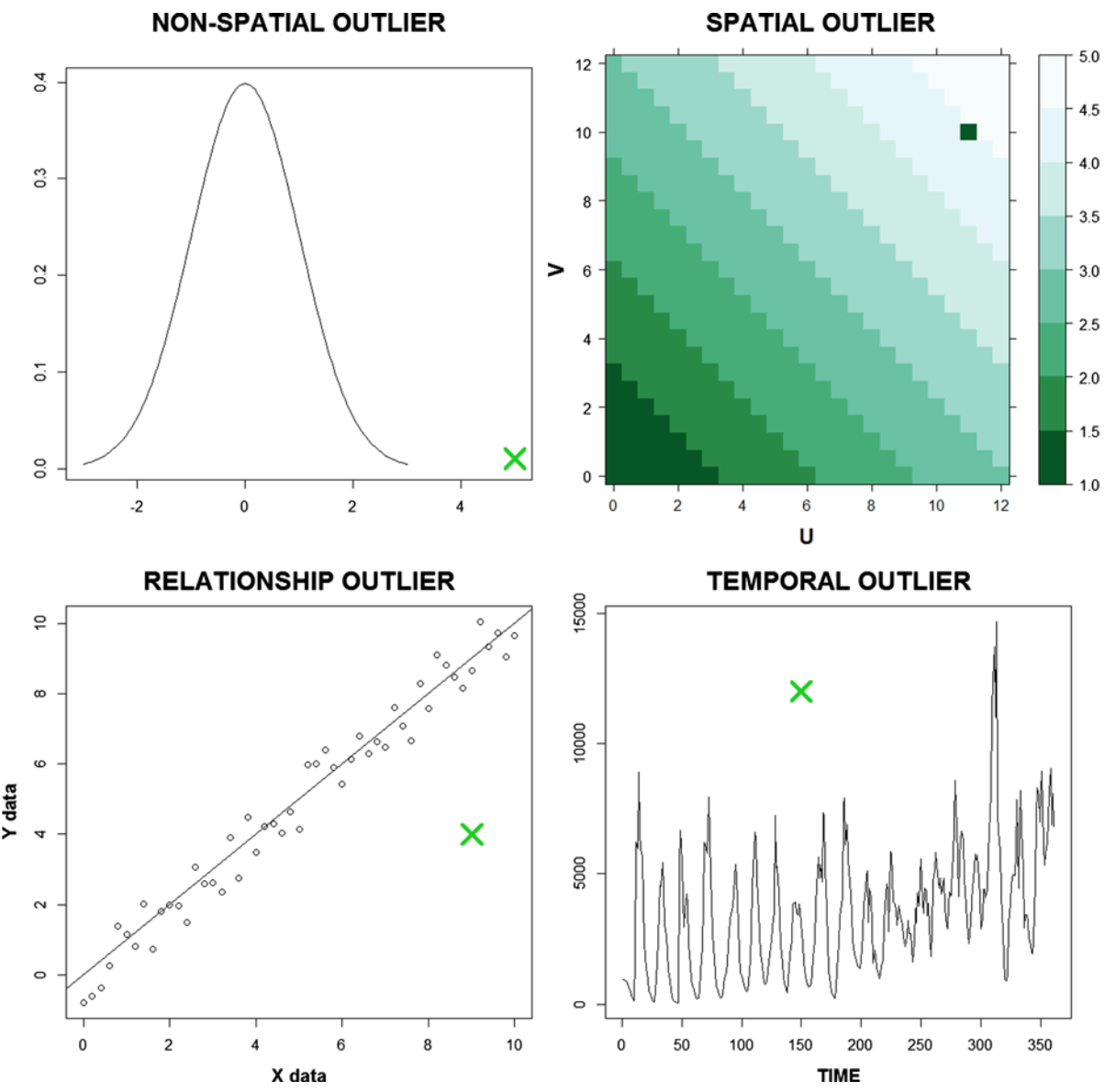}}
\caption{The evaluation of four different types of outliers in a geographical setting\cite{b7}}
\label{fig14}
\end{figure}

Figure \ref{fig14} shows non-spatial outliers, spatial outliers, relationship outliers and temporal outliers. Non-spatial outliers or univariate data can be simply removed using a boxplot and using inspection to remove the outliers. This is because outliers usually occur as a tail (right or left) in a sampled distribution.

In a bivariate case, the data considered in pairs. An outlier occurs when such a pair in an unusual in relation with respect to all the other data pairs and bagplots are used in this case treat this outlier. However, in the case of multivariate outlier detection, several different methodologies have been presented. One such is the detection with robust Mahalanobis Distances. This can be computed as follows:

\begin{equation}
MD_i = [(x_i - \mu)^\textbf{T} {\Sigma}^{-1} (x_i - \mu)]^{0.5}\ \text{for}\ i = 1,...,n \label{eq5}
\end{equation}

For low dimensionality data, Detection with Robust Principe Component Analysis (PCA) is considered. In this case, a standard linear algebra states for $p < n$, 

\begin{equation}
\textbf{LVL}^\textbf{T} = \textbf{R} \label{eq6}
\end{equation}

Where, V is the diagonal matrix of eigen values, L is the matrix of eigenvalues, and R is symmetric and positive definite.

And the SD (standard deviation) for a given component is:

\begin{equation}
SD_i = \sqrt[]{\sum^q_{k=1}\frac{t^2_{ik}}{v_k}} \label{eq7}
\end{equation}

The paper uses three geographically weighted (GWs) methods - One uses local Mahalanobis distances (MDs), whilst the other two, use outputs from a local principal component analysis (PCA). In conclusion, all the techniques perform adequately in an empirical and heuristic study. Detection performance is both illustrated using maps and measured numerically.

\section{Inference}

There exist several different methods for evaluating outlier detection over a spatial set of data. This is because data can be multivariate, bivariate, univariate or symmetric and hence the analysis applied is different in each case. Statistical based approaches are sensitive to changes in temporal correlations and a sudden change in the data distribution helps detect outliers. But are not suitable for real time applications. \cite{b1} Furthermore, since this approach uses histograms which do not rely on data distribution, its only useful for univariate data. Furthermore, computational cost for multivariate data is high if it is to be considered \cite{b10}. The nearest neighbor algorithm approach is unsupervised and does not make presumptions with any underlying distribution of data. Furthermore, the technique is fairly simple to apply \cite{b18} and only requires the selection of appropriate values for data cleaners \cite{b11}. However, computing the intra-distance among patterns is expensive and makes scalability of the algorithm difficult. Threshold values could lead to resulting the data being considered as false negative or false positive for outlier detection as selecting this value is a tough estimate. Inter cluster analysis is difficult and might result in elimination of effective data during spatial autocorrelation \cite{b12}. Artificial Intelligence based approaches are able to ``generalize from limited, noisy and incomplete data” \cite{b1}. Spatial and temporal semantics make the rules and fuzzy logic more complex and harder to implement and could poses a load on the memory.  Lastly, clustering based approaches are easily adaptable and do not tend to be supervised. But is again computational expensive due to intra-cluster distances among points.

\section*{Acknowledgment}

The author, Jacob John would like to thank Dr. Archana T for her continuous support throughout this paper. I would also like to thank Vellore Institute of Technology for their aid without which this paper wouldn't have been completed.

\end{document}